\documentclass[letterpaper, 10 pt, conference]{ieeeconf}  

\IEEEoverridecommandlockouts                              

\usepackage{graphicx} 
\usepackage{amsmath} 
\usepackage{amssymb}  
\usepackage{algorithm}
\usepackage{algpseudocode}
\usepackage{xcolor}
\usepackage{lipsum}
\usepackage{balance}
\DeclareMathOperator*{\argmin}{arg\,min}

\title{\LARGE \bf
Suture Thread Spline Reconstruction from Endoscopic Images for Robotic Surgery with Reliability-driven Keypoint Detection
}

\author{Neelay Joglekar$^{1}$, Fei Liu$^{1}$, Ryan Orosco$^{2}$, Michael Yip$^{1}$
\thanks{*This project was funded by the US Army Telemedicine and Advanced Technologies Research Center (TATRC) and NSF CAREER award \#2045803. N. Joglekar was supported by the UCSD Ledell Family Research Scholarship for Science and Engineering.}
\thanks{$^{1}$ Department of Electrical and Computer Engineering, University of California San Diego, La Jolla, CA 92093, USA {\tt\small \{njogleka, f4liu, yip\}@ucsd.edu}}
\thanks{$^{2}$ Department of Surgery, University of New Mexico, Albuquerque, NM 87131, USA {\tt\small RKOrosco@salud.unm.edu}}
}

\begin{document}
\bstctlcite{bstctl:nodash}

\maketitle
\thispagestyle{empty}
\pagestyle{empty}

\begin{abstract}

Automating the process of manipulating and delivering sutures during robotic surgery is a prominent problem at the frontier of surgical robotics, as automating this task can significantly reduce surgeons' fatigue during tele-operated surgery and allow them to spend more time addressing higher-level clinical decision making. Accomplishing autonomous suturing and suture manipulation in the real world requires accurate suture thread localization and reconstruction, the process of creating a 3D shape representation of suture thread from 2D stereo camera surgical image pairs. This is a very challenging problem due to how limited pixel information is available for the threads, as well as their sensitivity to lighting and specular reflection. We present a suture thread reconstruction work that uses reliable keypoints and a Minimum Variation Spline (MVS) smoothing optimization to construct a 3D centerline from a segmented surgical image pair. This method is comparable to previous suture thread reconstruction works, with the possible benefit of increased accuracy of grasping point estimation. Our code and datasets will be available at: https://github.com/ucsdarclab/thread-reconstruction.

\end{abstract}

\section{INTRODUCTION}

\begin{figure}[tbp]
    \centering
    \includegraphics[width=0.88\linewidth]{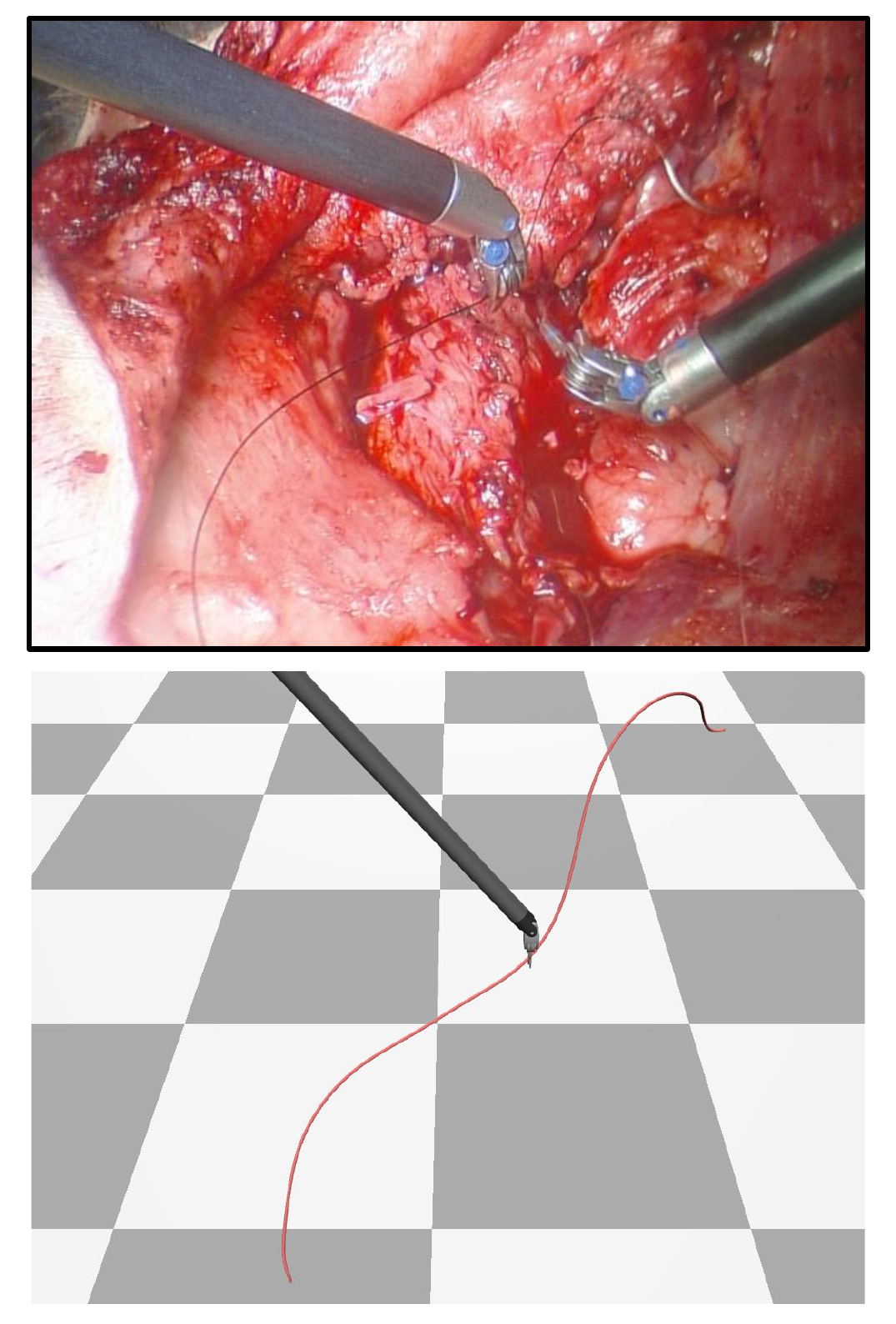}
    \caption{An example of our method's 3D suture thread reconstruction result from a real surgical image. The suture thread is difficult to see in the RGB image. Nonetheless, our method is able to pick out the features of the suture thread by using keypoints with an associated reliability metric, which can then be used to formulate and solve a spline smoothing problem. The result is an accurate thread reconstruction that is necessary for successful robotic suture thread manipulation.}
    \label{title}
    \vspace{-8mm}
\end{figure}

The promise of robotic surgery today is in teleoperation, where the robot can augment a surgeon's performance by scaling down motions, removing tremor, and even allowing for remote operations across short distances. However, future applications look beyond teloperation and towards the automation of surgical tasks, which is needed to address the impending doctor shortage. One of the most prevalent surgical tasks then to automate involves robot manipulation and automation of suture placement. In particular, suture thread manipulation involves thread tracking and performing complex suturing and knot-tying operations. Since surgeons need to use suture thread often to close incisions, automating thread manipulation shall significantly reduce a surgeon's workload and fatigue.
    
One important subtask during suturing is suture thread reconstruction. This is the process of taking 2D surgical images and constructing a 3D centerline of the suture thread in the surgical scene. This 3D centerline is necessary for finding grasping points along the thread and planning the placement and movement of instruments during knot tying such that entanglement does not occur.
    
Although many robust methods have been developed for general object reconstruction from images, suture thread reconstruction is particularly difficult. Suture thread is incredibly thin; the width of features in the image, based on the common optics for endoscopic robotic platforms, occupy less than 5 pixels, or about 0.25\% of the image width. Most object reconstruction methods in computer vision struggle with reconstructing thin features. Furthermore, given the very low signal-to-noise (SNR) in endoscopic images, which are generally tend very noisy and lighting-sensitive, 3D reconstructions will also tend to exhibit noise. Without any metric to discern the reliability of a stereo triangulation, an algorithm for reconstruction may place false confidence in its results, which then can easily lead to the failure of completing a non-trivial manipulation task such as suturing. Previous suture thread reconstruction methods do not explicitly check this pixel reliability.


An additional step in this reconstruction process involves generating a continuous, smooth centerline. When solving the visual-servoing control problem, one is usually required to convert reconstructions to a spatially smooth model such that sampling can be done in a continuous space and gradients may be taken. Furthermore, smooth curves are more realistic representations of suture thread, and therefore are likely more accurate. Hence, a suture thread reconstruction should be represented by a smooth, continuous, and well-parameterized curve.
    
In this paper, we introduce a suture thread reconstruction \textit{and} spline model fitting method that uses reliability-based keypoints to define and solve a Minimum Variation Spline (MVS) problem, the result of which represents the thread's 3D centerline. The ordered keypoints anchor the reconstructed spline curve to reliable features in the given surgical images, which prevents noisy features from influencing the result. The MVS optimization ensures the curve shape is smooth and realistic, imitating the kinematic bending properties of real suture thread.
    

\section{RELATED WORKS}

In robotic surgery, the primary visual feedback for 3D navigation and localization of objects is from stereo-cameras. 3D reconstructions therefore rely on good stereo matching. Stereo matching involves matching blocks of pixels in the left and right images of a stereo pair, based on similarity, smoothing, and etc., and then assigning depths based on the relative locations of the matched blocks. Many of these methods have been implemented in OpenCV, a popular computer vision library \cite{bradski2000}. Stereo matching in the case of suture threads comes with considerable challenges: the feature widths of the suture thread are extremely small, the camera disparity of stereo cameras is necessarily very tight (typically 8mm or less), and the lighting conditions are especially poor. This results in noisy images and also noisy reconstructions. An interesting set of methods that address such challenging scenes are presented in \cite{kim2015} and \cite{geiger2011}. They augment basic stereo matching with reliability metrics, which measure the likelihood that a pixel's assigned depth is correct. These metrics are used to anchor reconstructions at the most reliable image features. Unfortunately, to our knowledge, no such stereo matching method has been successfully applied to reconstruct a thin structure, such as suture thread.
    
For suture thread reconstruction, a few works exist. The method presented in \cite{lu2021} constructs a 3D graph by matching pixels with similar ordering values and then designates the lowest cost path in this graph as the 3D centerline. The method presented in \cite{jackson2018} uses explicit stereo matching and implicit ordering-based matching to initialize a smooth 3D spline. Both methods perform respectably well, but their accuracy can be improved. In addition, both methods require the user to manually specify an initial point in the thread images, which is not suitable for autonomous procedures and for the dynamic nature of surgery.
    
Curve generation, the process of constructing full curves that satisfy boundary conditions, includes many methods designed to optimize curve smoothness. Previous work has explored the Minimum Variation Curve (MVC) smoothing problem, an optimization problem that minimizes the variation in curvature along a curve while satisfying given constraints \cite{moreton1992}. The work presented in \cite{moll2006} demonstrates that the MVC objective function is similar to the internal strain equation of wire-like structures, and therefore solving MVC can produce realistic 3D curves. The Minimum Variation Spline problem presented in \cite{berglund2003} is a modification of MVC that specifically smooths splines instead of general curves. Again, to our knowledge, no such method has been applied to suture thread reconstruction from images.
    
Unlike prior thread reconstruction methods, the method presented here solves the following problems: Performing suture thread reconstruction without user input (e.g., a seed point), explicitly recognizing reliable thread image features, developing keypoints via clustering, and explicitly minimizing the smoothing energy of the output curve.

\section{METHOD}

Our approach is to perform spline-based reconstruction that is anchored to highly reliable points in 3D space and minimizes a energy term associated with the smoothness of a curve. 
Our method can be decomposed into 3 steps: stereo-matching preprocessed images, choosing reliable, ordered keypoints from the stereo matching result, and using these keypoints to construct and solve an MVS optimization problem. The result of this method is a smooth 3D spline representation of the thread centerline.

\subsection{Stereo Matching Reconstruction on Preprocessed Images}

\subsubsection{Image Preprocessing}

The input left and right pair of raw surgical images (converted to greyscale) are denoted as $I_L^{raw}, I_R^{raw} : \mathbb{Z}^+ \times \mathbb{Z}^+ \to \mathbb{R}^+$. (i.e. we define images as mappings from 2D discrete pixels, denoted by $p$, to real-number intensities. Similar notation is used throughout this paper.) $I_L^{raw}$ and $I_R^{raw}$ are initially stereo-rectified.  This makes epipolar lines in the raw image pair horizontal, a common preprocessing step before stereo matching. 

The rectified images $I_L^{rec}, I_R^{rec}$ are then segmented and ``lifted". Lifting images involves using a mask to select only the segmented pixels in each image, $P_L, P_R \subset \mathbb{Z}^+ \times \mathbb{Z}^+$, and setting all remaining pixels to a background value:
$$I_L^{lift}(p) =
        \begin{cases}
            I_L^{rec}(p) & p \in P_L\\
            255 & \textrm{otherwise}
        \end{cases} \eqno{(1)}$$
\vspace{-4mm}

\noindent $I_R^{lift}$ can be similarly computed from $P_R$. Note that there isn't a specific method required for segmentation; the most modern technique may be used. We used color segmentation in our simulated experiments and hand segmentation in real evaluations just to demonstrate.

\subsubsection{Stereo Matching on Lifted Images}

Next, we perform stereo matching on $I_L^{lift}$ and $I_R^{lift}$.
Each $p$ is assigned an integer pixel disparity $d \in [0, \alpha]$ by performing the following energy minimization:
$$ E(d, p) = \sum_{p' \in g_p} \left(I_L^{lift}(p') - I_R^{lift}((p'_x - d, p'_y))\right)^2 \eqno{(2)}$$ 
\vspace{-3mm}
$$d = \argmin_{d_i \in [0, \alpha]} E(d_i, p) \eqno{(3)}$$
Here $g_p \subset P_L$ is a set of segmented pixels surrounding $p$, $(\cdot)_x$ and $(\cdot)_y$ denote a pixel's $x$ and $y$ coordinates, and $\alpha$ is a maximum disparity threshold (typically around 80). It should be noted that stereo matching on lifted images reduces noise by preventing any pixels from being matched to the unsegmented background of the right image.
    
Once all disparities are calculated, an initial thread reconstruction is done using OpenCV's \textit{reprojectImageTo3D} function \cite{bradski2000}, in the form of a depth map $D : P_L \to \mathbb{R}$.

\vspace{-3mm} 

$$[X_p, Y_p, Z_p, W_p]^T = \boldsymbol{Q} [p_x, p_y, d_p, 1]^T \eqno{(4)}$$

\vspace{-5mm}

$$D(p) = Z_p/W_p \eqno{(5)}$$

\noindent where $\boldsymbol{Q}$ is the disparity-to-depth matrix and other unnamed values are intermediate variables.

\subsection{Reliability-Based Keypoint Selection from Initial Image-Based Reconstruction}

Next, we utilize pixel reliability to select and order keypoints from our depth map $D$. 
It should be noted how our use of keypoints makes our method unique compared to previous thread reconstruction works \cite{lu2021} and \cite{jackson2018}. For one, both works require manual input to extract ordering information, but our method does not. More importantly, although \cite{lu2021} does account for some uncertainty, neither method explicitly evaluates and accounts for pixel reliability while our method does.


\subsubsection{Reliability-Based Pruning}

$D$ is noisy and contains inaccurate segments. This is especially clear in areas where the thread tangent is aligned with the images' epipolar lines, or where multiple thread segments are horizontally close to each other. These would be difficult as well for human eyes to triangulate.

\begin{figure}[tbp]
    \centering
    \vspace{2mm}
    \includegraphics[width=0.9\linewidth]{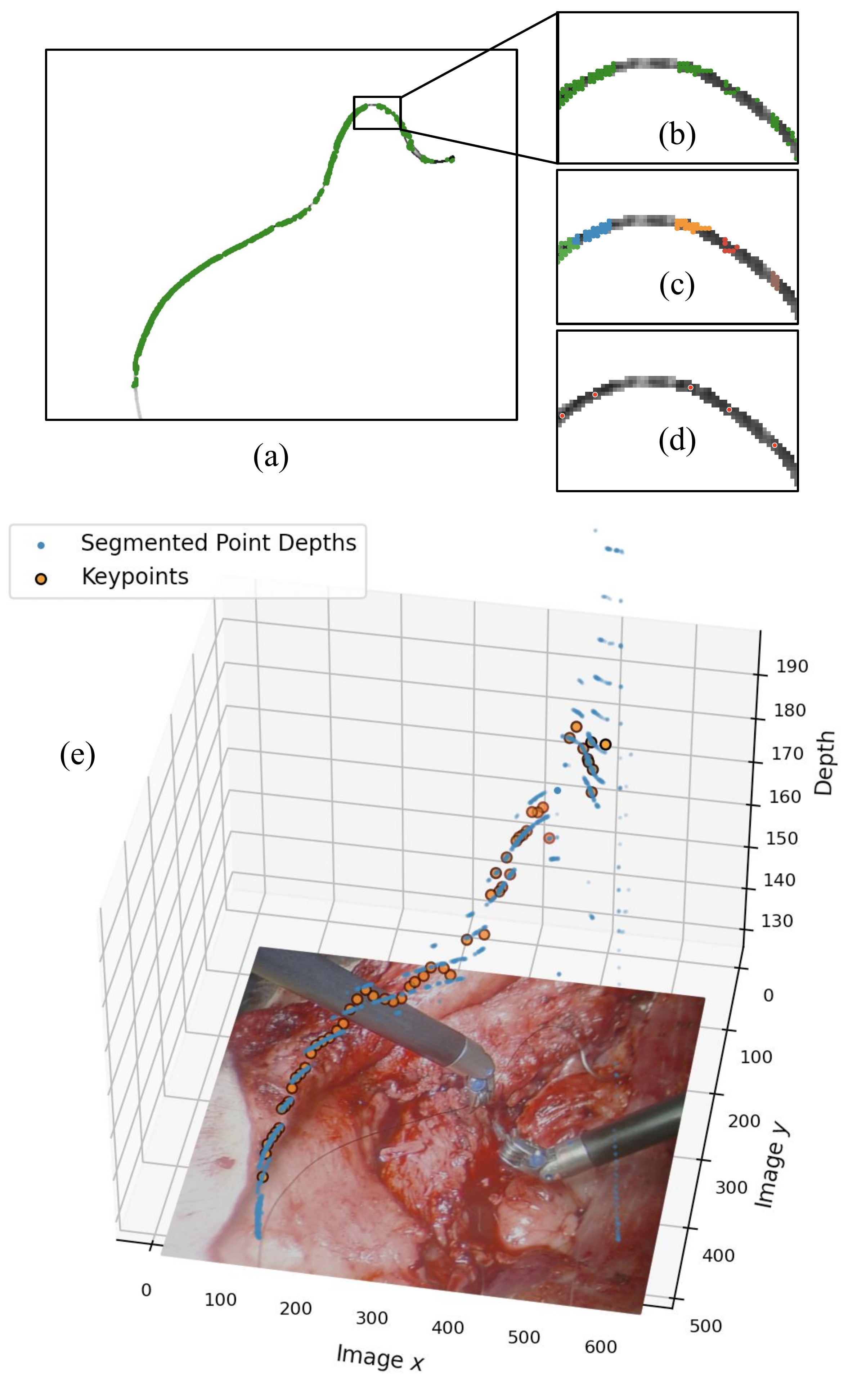}
    \vspace{-3mm}
    \caption{\textbf{The keypoint selection process}. (a) shows the set of reliable points (in green) over the entire image. (b) shows a zoomed in section of the reliable points. (c) These points are clustered based on locality in the left image. (d) The cluster centroids are designated as keypoints (in red). (e) The resulting keypoints avoid the most unreliable sections of the stereo matching point cloud, making them ideal anchor points for 3D curve fitting. (Image plane ticks are in pixels, and depth is in mm.)}
    \label{keypt_sel}
    \vspace{-5mm}
\end{figure}

To overcome this issue, we prune points from $D$ using a disparity reliability metric. As in \cite{kim2015} and \cite{geiger2011}, Each $p$ is assigned a reliability by comparing the best and second best disparities, $d_p^{min}$ and $d_p^{next}$, that minimize (2). The energies of these disparities are denoted as $E_p^{min}$ and $E_p^{next}$:
$$R(E_p^{min}, E_p^{next}) = \sigma\left(\epsilon_1 \left(\frac{E_p^{next} - E_p^{min}}{\epsilon_2 E_p^{min}} - \epsilon_3\right) \right) \eqno{(6)}$$
It should be noted that $d_p^{next}$ is chosen such that $\|d_p^{next} - d_p^{min}\| > 2$ as the disparities near $d_p^{min}$ are likely to have similar energies. The sigmoid function, $\sigma(\cdot)$, models a sharp switch from low to high reliability, and the function is scaled and shifted using $\epsilon_1, \epsilon_2,\textrm{ and }\epsilon_3$ (we set these values to 8, 5, and 0.8 respectively). We use a threshold to create a set of reliable pixels $P_L^R = \{p \in P_L : R(E_p^{min}, E_p^{next}) > 0.9\}$. This result is visualized in Fig. \ref{keypt_sel} (a, b).

\subsubsection{Keypoint Selection via Clustering}

$P_L^R$ is clustered using breadth-first search (BFS) to form keypoints. BFS is performed multiple times, each time starting at an arbitrary unexplored $p^R \in P_L^R$ and growing until the frontier is empty or a max size threshold is met. The neighbors of each $p^R$ are designated as all $p'^R$ \textit{within a manhattan distance of at most 2} of $p^R$ to allow clusters to grow beyond small gaps. If the explored set meets a minimum size threshold as well, then a cluster $c \subset \mathbb{R}^3$ is formed from the set of 3D points associated with the explored set. The set of clusters is denoted as $C$. This process stops once no more valid clusters can be formed. Lastly, each cluster centroid is designated as \textit{a keypoint}, resulting in the keypoint set $K \subset \mathbb{R}^3$.

This process is visualized in Fig. \ref{keypt_sel} (c, d, e). The resulting keypoints successfully avoid the most unreliable portions of $D$ while remaining well-distributed around the curve.

\subsubsection{Topological Ordering of Keypoints}

    
As a preprocessing step, the clusters are ``solidified" to increase the chance that each cluster spans the entire width of the thread. For each $c \in C$, all nearby $p \in P_L$ (typically within 1-2 pixels of $c$) are added to $c$.
    
Next, an adjacency graph $\mathcal{A}$ is constructed to identify the adjacency of keypoints in $K$. For each keypoint $k \in K$, its adjacent keypoints are defined by performing BFS on $P_L$. Starting from an arbitrary $p^c \in c$, where $c$ is the cluster corresponding to $k$, BFS (where, unlike before, edges are defined by a pixel's 8 canonical neighbors) is performed with the modification that \textit{no pixels in clusters other than} $c$ can be placed in the frontier. For every cluster $c'$ encountered during this operation, an edge between $k$ and the associated $k'$ is added to $\mathcal{A}$.
    
A modified depth-first search (DFS) is then performed on $\mathcal{A}$ to determine the topological keypoint ordering $\phi : K \to \mathbb{Z}^+$. Starting at a keypoint with exactly 1 edge, DFS is performed such that only \textit{the closest unexplored adjacent keypoint} is placed in the frontier at each iteration. As each keypoint is explored, it is mapped to the next available index in $\phi$. This modified DFS is robust against undirected cycles in $\mathcal{A}$ formed when clusters don't fully span the width of the thread (hence the necessity of cluster solidification).
    
\begin{figure}[tbp]
    \centering
    \vspace{3mm}
    \includegraphics[width=\linewidth]{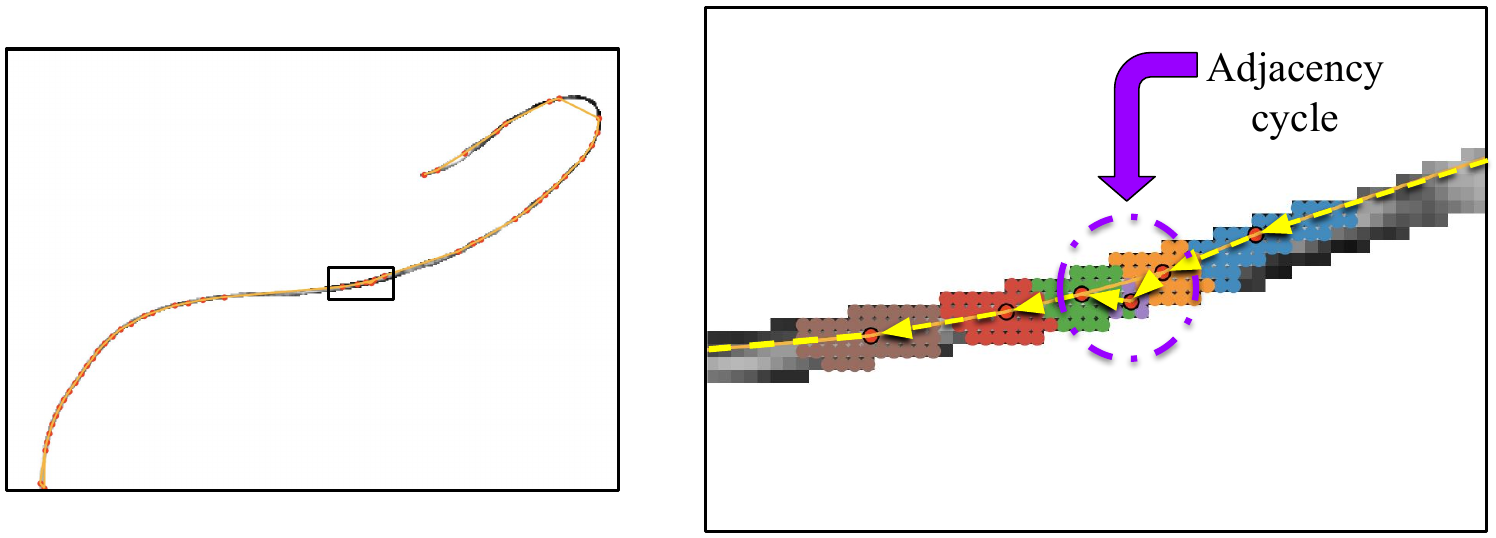}
    \caption{A visualization of the solidified clusters (distinguished by color), adjacency graph $\mathcal{A}$ (consisting of the red keypoints and orange edges between them), and the ordering result (yellow dashed arrows) in a section of thread. This ordering method is robust to cycles in $\mathcal{A}$.}
    \label{ordering}
    \vspace{-3mm}
\end{figure}

To ensure that the first and last keypoints line up with the real thread endpoints, a final BFS is performed at each of these 2 keypoints. This BFS is performed on all \textit{unclustered} $p$ reachable from each associated $c$ (with reachability defined the same as in the previous BFS). If a significant number of these pixels are not reachable from any neighboring $c'$, the pixel farthest from $k$ is converted to a keypoint with the same depth as $k$. $\phi$ is also updated as required. The entire ordering process is visualized in Fig. \ref{ordering}. The result is an ordered set of keypoints in 3D space.

\subsection{Recovering a Minimum Variation Spline Model with Keypoint-Derived Boundary Conditions}


To use a reconstruction in any nontrivialized control problem, a continuous function describing the reconstructed thread centerline is desirable so that it may be sampled in a continuous-space fashion or such that a gradient may be taken. As a result, it's advantageous to fit a spline to the previously computed keypoints.
    
\begin{figure}[tbp]
    \centering
    \vspace{3mm}
    \includegraphics[width=0.95\linewidth]{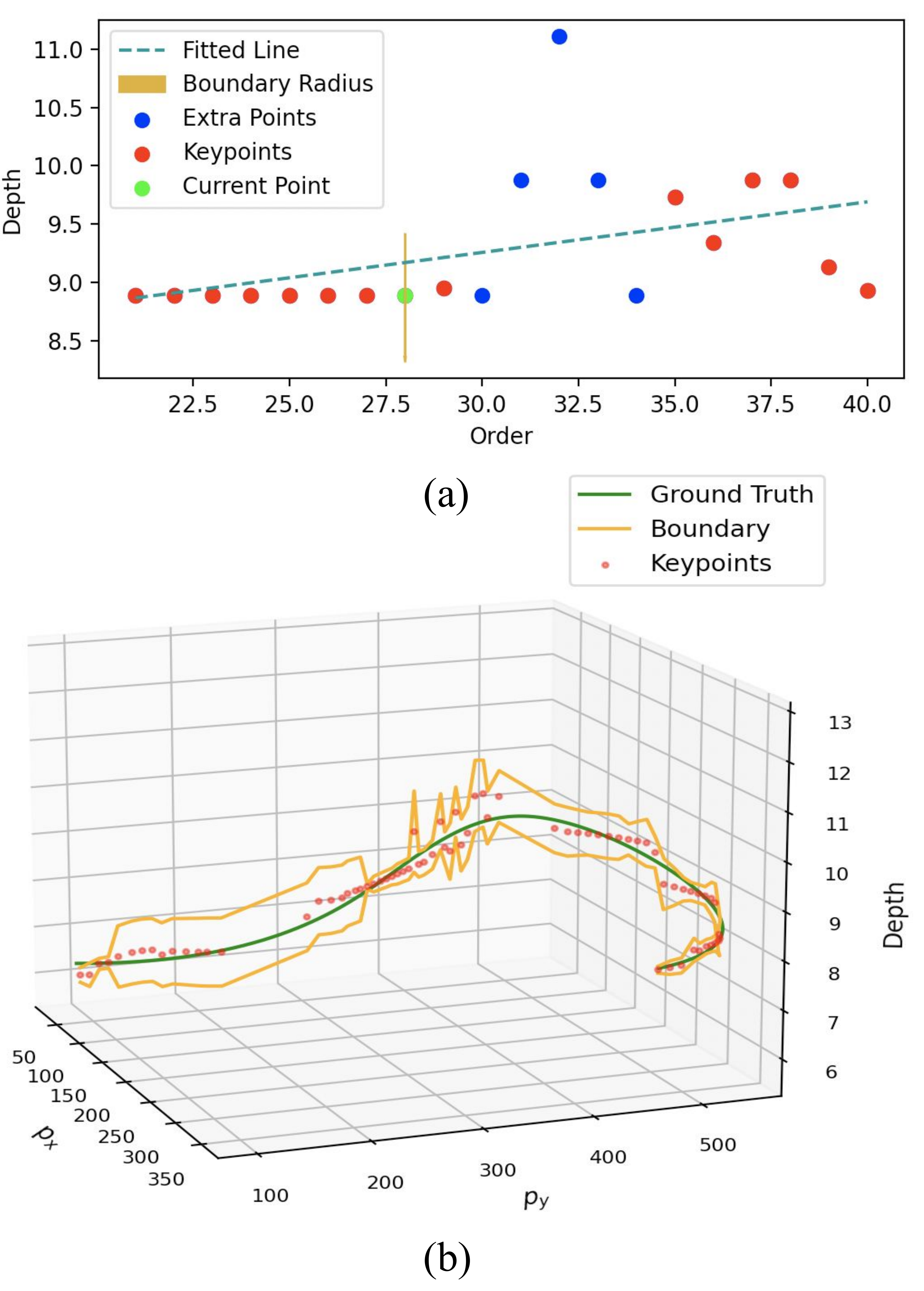}
    \vspace{-3.5mm}
    \caption{(a) Boundaries for the Minimum Variation Spline problem are constructed based on lines fitted to each local group of keypoints. (b) The constructed boundaries tend to include as much of the ground truth curve as possible without being too wide. ($p_x$ and $p_y$ are pixel coordinates in the 2D image plane, and depth is in cm. The units of ``order" are the indicies of each point in ordering $\phi^H$.)}
    \label{boundaries}
    \vspace{-5mm}
\end{figure}

In addition, to best match reality, a reconstructed thread centerline should mostly capture the thread's kinematic deformation properties. Hence, simply initializing a spline that best fits $K$ ignores the physical properties of real suture thread and can lead to unrealistic fits. In addition, as shown in Fig. \ref{keypt_sel}(e) and Fig. \ref{boundaries}, the set of keypoints are still too noisy to be simply strung together. To account for these factors, we instead use the keypoints to define the constraints and initial guess for an MVS problem, from which a smooth spline can be constructed.
    
As a preprocessing step, extra points are selected between keypoints in regions where keypoints are sparse. Each pair of keypoints is evaluated by checking the number of segmented pixels between them. If this number is too large, extra points are chosen and ordered based on their location between these keypoints. The resulting point set $H \supset K$ and ordering $\phi^H$ includes extra information in unreliable regions of $D$ that is otherwise ignored by $K$.

    
Next, both $K$ and $H$ are used to define depth constraints for the MVS optimization. For each ordered keypoint $k_i$, the depths of all points in $H$ between $k_{i-r_k}$ and $k_{i+r_k}$ ($r_k$ is typically $|K| / 10$) are collected and fitted with a least squares line $L_{k_i}$. The lower and upper depth boundaries, $\underline{K}, \overline{K} : H \to \mathbb{R}$, at $k_i$ are then defined as follows:
$$\underline{K}(k_i) = k_{i, z} - 1.5\| L_{k_i}(k_i) - k_{i, z} \| $$$$\overline{K}(k_i) = k_{i, z} + 1.5\| L_{k_i}(k_i) - k_{i, z} \| \eqno{(7)}$$
Here, the $(\cdot)_z$ subscript denotes the depth component of the variable in consideration. The boundaries between keypoints are linearly interpolated, and any boundaries that are too restrictive are widened based on a threshold. A visualization of these boundaries is shown in Fig. \ref{boundaries}.

As a final step before optimization, The spline $S : \mathbb{R} \to \mathbb{R}^3$ with parameterization $u$ is initialized. A set of initial points $H^i$ is constructed from $H$ by modifying all point depths to match the centerline between $\underline{K}$ and $\overline{K}$. Next, using SciPy's \textit{splprep} function \cite{dierckx1982} \cite{dierckx1996} \cite{virtanen2020}, $S$ is initialized from $H^i$, a uniform knot sequence, a degree of 4, and $m = 15$ control points, $b \in \mathbb{R}^{3 \times m}$.
    
Finally we solve the following MVS problem, inspired by \cite{berglund2003}, to smooth $S$ via its control points:
    $$\min_{b_z \in \mathbb{R}^m} \int_{0}^{\ell} \frac{
    \left(\frac{\partial}{\partial u} \kappa (b_z, u)\right)^2
    }{
    \sqrt{1 + \left(\frac{\partial}{\partial u} S_z(b_z, u)\right)^2}
    } du \eqno{(8)}$$
    such that
    \begin{align*}
        \underline{K} \leq S_z(b_z, u) &\leq \overline{K} & u \in [0, \ell]\\
        \frac{\partial^j}{\partial u^j} S_z(b_z, 0) &= \frac{\partial^j}{\partial u^j} L_{k_1}(k_1)\\
        \frac{\partial^j}{\partial u^j} S_z(b_z, \ell) &= \frac{\partial^j}{\partial u^j} L_{k_{n}}(k_{n}) & j = 0, 1
    \end{align*}
 $\kappa$ is the curvature of $S_z$, defined as in \cite{berglund2003}, while $\ell$ is the last parameter value of $S$. In addition to the depth constraint enforced by $\underline{K}$ and $\overline{K}$, the endpoint positions and derivatives of $S_z$ are fixed to their previously-computed least squares lines at the endpoint keypoints, $k_1, k_n$. Any sufficient optimizer can be used to solve this problem.
Once generated, the final spline is transformed into the camera coordinate frame. This is done by transforming the control points:
$$b^{cam} = 
\begin{bmatrix}
    b_z & b_z & b_z
\end{bmatrix}^T
\odot \left(\boldsymbol{G}^{-1} 
\begin{bmatrix}
    b_x & b_y & \boldsymbol{1}
\end{bmatrix}^T \right) \eqno{(9)}$$

\noindent where $\odot$ is the Hadamard Product and $\boldsymbol{G}$ is the camera matrix.

\section{EXPERIMENTS \& RESULTS}

\begin{figure*}[tbp]
    \centering
    \includegraphics[width=\linewidth]{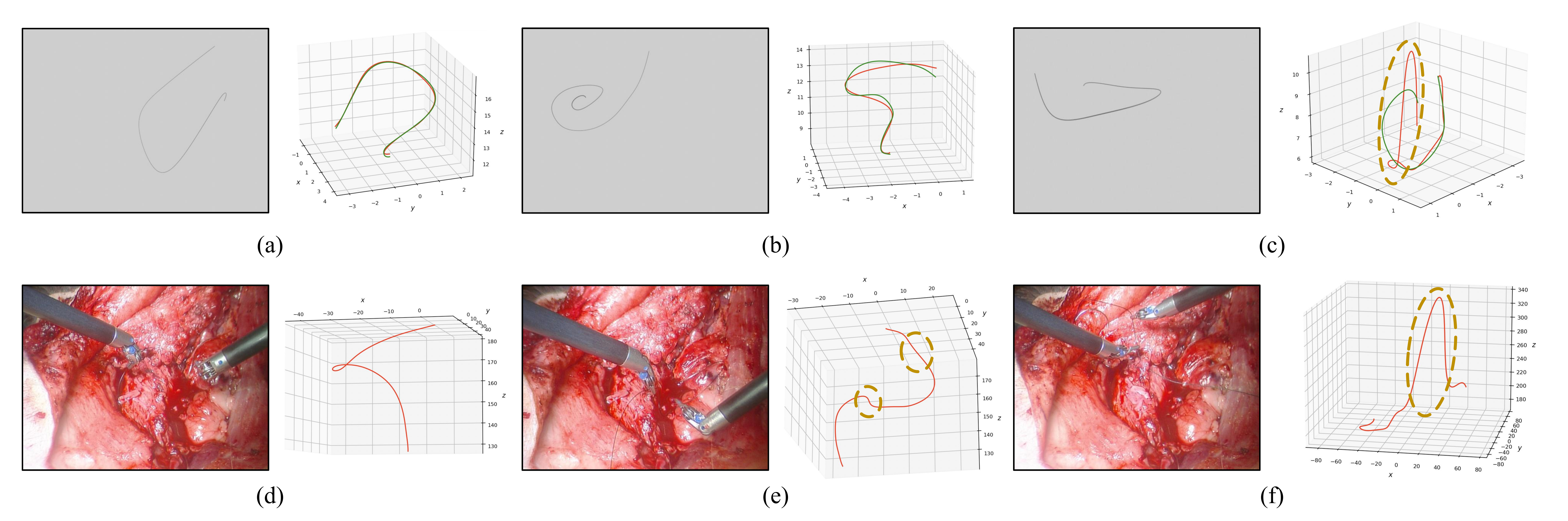}
    \caption{Reconstruction results (in red) from image pairs in (a,b,c) simulated and (d,e,f) real datasets are visualized. In the simulated results, ground truth curves (in green) are also displayed. (a) and (d) are very accurate reconstructions, (b) and (e) are less accurate but still successful, and (c) and (f) are reconstruction failures. Areas of questionable accuracy are circled. Most reconstructions produced from the from both the simulated and real datasets were accurate. (The units in the simulated dataset results are cm while those in the real dataset results are in mm.)}
    \label{examples}
\end{figure*}

We built our own datasets of simulated thread images and real-life surgical images to test our method. To our knowledge, no datasets have been released from either \cite{lu2021} or \cite{jackson2018}, and thus the inclusion of real-life surgical images is especially important in this work to show the efficacy and failures of current approaches. The simulated dataset was built from stereo images and ground truth data collected from 3D models in Blender. For the real dataset, we performed live surgical experiments on a pig with a daVinci Research Kit \cite{richter2021bench}. Then, to the best of our ability, we compare our method to previous suture thread reconstruction methods presented in Lu et al. \cite{lu2021} and Jackson et al. \cite{jackson2018}.

All code was implemented in Python and run on a 2 GHz Quad-Core Intel Core i5 computer. Image rectification and deprojection was implemented using OpenCV's \textit{stereoRectify} and \textit{reprojectImageTo3D} functions \cite{bradski2000}. To perform spline initialization and solve MVS we used SciPy's \textit{splprep} function and SLSQP optimizer \cite{dierckx1982} \cite{dierckx1996}  \cite{virtanen2020} \cite{kraft1988}. It should be noted that better optimizers are likely available, but SLSQP was relatively convenient to use. Our code and datasets will be available at: https://github.com/ucsdarclab/thread-reconstruction.

\subsection{Evaluation on Simulated Suture Thread Dataset}


Our synthetic thread dataset was constructed from Blender NURBS models that mimicked real thread shapes. The data consists of stereo image pairs with resolution $640 \times 480$ as input and each model's NURBS control points as ground truth. These images are also easily color segmented due to the lack of a surgical background. 10 separate curves were modeled and photographed in 4 different orientations, resulting in 40 total data points. Each curve is 0.1 mm wide and between 60-140 mm long.
    
We evaluated our method with 3 quantitative measurements: mean curve error, $e_S$, max curve error, $e_S^{max}$, and curve length error, $e_{\ell}$. $e_S$ and $e_S^{max}$ measure the mean and maximum distance, respectively, between any point on the reconstructed curve and its closest point on the ground truth. $e_{\ell}$ is the absolute value of the length difference between the reconstructed curve and the ground truth. The curve errors are a better measure of accuracy than curve length error, and we track curve length error mainly for comparison purposes.
    
The evaluation results of our method are summarized in Table \ref{sim_Results}. Our $e_S$ is impressively low, averaging at just 1.2 mm. Our $e_S^{max}$ is larger on average, but qualitative analysis of our results reveals that reconstructions with $e_S^{max} \geq 5$ mm are often still accurate except in some small curve segment. The advantage of having a reliability metric is evident here, as then we can use this to avoid grasping at these locations. Of the 40 tests, 5 resulted in reconstruction failure (not factored into Table \ref{sim_Results}). However, 3 of these were optimization failures that could likely be fixed with a better optimizer.

\begin{table}[tbp]
    \centering
    \vspace{3mm}
    \caption{Mean and standard deviation of reconstruction curve error ($e_S$), max curve error ($e_S^{max}$), and length error ($e_L$) on simulated dataset (all measurements in mm).}
    \label{sim_Results}
    \begin{tabular}{|c|c|c|c|c|c|}
        \hline
        $\mu(e_S)$ & $\sigma(e_S)$ & $\mu(e_S^{max})$ & $\sigma(e_S^{max})$ & $\mu(e_{\ell})$ & $\sigma(e_{\ell})$\\
        \hline
        \hline
        1.2 & 0.75 & 6.2 & 5.1 & 7.7 & 8.6\\
        \hline
    \end{tabular}
    \vspace{-5mm}
\end{table}

Our results indicate that our method seems to improve upon some aspects of \cite{lu2021} and \cite{jackson2018}. \cite{lu2021} uses a ``grasping error" metric, which is similar to curve error except only evaluated at 1 grasping point. Their grasping error tends to be much larger than our curve errors, with a mean of 7.2 mm and max of 11.0 mm on average, but it should be noted that this metric was taken from real data instead of simulated data. \cite{jackson2018} reports RMS and max curve errors of 1.4 mm and 4.0 mm on average, respectively, both values comparable to ours; however, these values were taken from experiments on a small, simple dataset, often with little depth variation. Both \cite{lu2021} and \cite{jackson2018} report better curve length errors of 1.3 mm and 6.0 mm on average, respectively, but this metric is unimportant when performing suture thread manipulation. It is still difficult to determine the accuracy of these comparisons because they use completely different datasets, but our method is at least comparable and may provide a more accurate sense of grasping \textit{regions}.

\subsection{Evaluation on Real Surgical Dataset}


Our real surgical image dataset consisted of ten $640 \times 480$ surgical image pairs taken by a dVRK stereo camera~\cite{richter2021bench}. 
We evaluated our method on this dataset using mean and max 2D reprojection errors, $e_{2D}^{mean}$ and $e_{2D}^{max}$, on the left and right images and a quality rating. After projecting the final spline onto the stereo images, the reprojection error is computed by measuring the pixel distance between each projected point to the closest segmented pixel. The quality rating is a manual scoring on a scale from 1 (very bad) to 5 (very good), based on how plausible the final spline looks. Reconstructions with ratings below 3 are designated as failures.
    
The evaluation results are displayed in Table \ref{real_Results}. The reprojection errors for all tests are quite small. In addition, 8 out of the 10 reconstructions were given a quality rating of at least 3, most being 5's, because these curves looked smooth and plausibly accurate. Only two reconstructions were significantly noisy and implausible, resulting in ratings below 3. These results seem similar to those from the simulated dataset, despite increased image noise. Examples of both good and bad quality curves are visualized in Fig. \ref{examples}.

\begin{table}[tbp]
    \centering
    \vspace{2mm}
    \caption{2D reprojection error ($e_{2D}$, in pixels) and qualitative analysis on a scale of 1 to 5 of real dataset.}
    \label{real_Results}
    \begin{tabular}{|c|c|c|c|c|c|}
        \hline
        Thread \# & \multicolumn{2}{|c|}{Left Image} & \multicolumn{2}{|c|}{Right Image} & Quality\\
        & $e_{2D}^{mean}$ & $e_{2D}^{max}$ & $e_{2D}^{mean}$ & $e_{2D}^{max}$ &\\
        \hline
        \hline
        1 & 0.43 & 2.04 & 0.48 & 2.24 & 3\\
        2 & 0.96 & 6.53 & 0.98 & 7.55 & 2\\
        3 & 0.39 & 0.83 & 0.37 & 0.70 & 5\\
        4 & 0.37 & 0.67 & 0.42 & 1.90 & 5\\
        5 & 0.39 & 0.68 & 0.43 & 6.09 & 5\\
        6 & 0.72 & 4.33 & 1.12 & 7.76 & 4\\
        7 & 0.51 & 3.71 & 0.61 & 6.49 & 5\\
        8 & 0.51 & 4.28 & 0.65 & 5.54 & 5\\
        9 & 0.50 & 4.29 & 0.62 & 7.48 & 2\\
        10 & 0.39 & 1.10 & 0.40 & 0.99 & 4\\
        \hline
    \end{tabular}
    \vspace{-2mm}
\end{table}

Unfortunately, it is difficult to perform a fair comparison with other works \cite{lu2021, jackson2018}. Jackson et al.~\cite{jackson2018} uses colored paper to imitate a surgical background, which is much less noisy than real surgical data. Lu et al.~\cite{lu2021} does use images from experiments on real tissue and, as mentioned earlier, reports relatively impressive grasping error numbers. However, since grasping error is evaluated at only 1 point on the reconstruction, it is difficult to tell how accurate their entire reconstruction is.

\section{CONCLUSIONS \& FUTURE WORK}

We present and test a method to construct a smooth 3D spline representation of suture thread using reliable keypoints derived from a stereo matching reconstruction. We verify the accuracy of our method on simulated and real datasets, achieving a mean curve error of just 1.2 mm and successfully reconstructing 80\% of our real surgical images.
    
The main drawback of this work is its susceptibility to reconstruction failure in certain scenarios, as shown in Fig. \ref{keypt_sel} (c, f). These are very challenging figures, however, which may be affecting the reliability of large groups of keypoints or the accuracy in ordering. In addition, this method relies on multiple hand-tuned threshold values and ignores self-intersection/occlusions, which may reduce its generalizability.
    
Regardless of the current limitations, this new formulation provides an interesting new set of options towards the task of autonomous suturing that were not possible before.
This method is easier to use because it requires no manual user input.
In addition to aiding reconstruction, our reliability metric can likely help identify the most accurate thread grasping points. Lastly, our smoothing optimization better ensures the physical accuracy of the reconstruction result. Ultimately, this work moves forward with the robot's understanding and evaluation of reliability of reconstructions for suture thread, which will be a critical enabler to autonomous suturing in real-world, unstructured scenarios.
    

\addtolength{\textheight}{-12cm}   





\clearpage
\balance
\bibliographystyle{IEEEtran}
\bibliography{fulldraft}

\end{document}